\documentclass[10pt, conference, compsocconf]{IEEEtran}

\usepackage{times}
\usepackage{epsfig}
\usepackage{graphicx}
\usepackage{amssymb}
\usepackage{epstopdf}
\usepackage{array}

\usepackage{multirow}
\usepackage{algorithm}
\usepackage{algorithmic}
\usepackage{wrapfig, blindtext}
\usepackage{amsmath}
\DeclareMathOperator*{\argmax}{arg\,max}

\begin{document}

%
\title{The GIST and RIST of Iterative Self-Training for Semi-Supervised Segmentation}

\author{Eu Wern Teh\textsuperscript{1,2,3}
	\and
	Terrance DeVries\textsuperscript{1,2}
	\and
	Brendan Duke\textsuperscript{3,4}
	\and
	Ruowei Jiang\textsuperscript{3}
	\and
	Parham Aarabi\textsuperscript{3,4}
	\and
	Graham W.~Taylor\textsuperscript{1,2}
	\and
	\hspace{3.5em} \textsuperscript{1}University of Guelph\qquad
	\textsuperscript{2}Vector Institute\qquad
	\textsuperscript{3}Modiface, Inc.\qquad
    \textsuperscript{4}University of Toronto
}

\maketitle

\begin{abstract}
We consider the task of semi-supervised semantic segmentation, where we aim to produce pixel-wise semantic object masks given only a small number of human-labeled training examples. We focus on iterative self-training methods in which we explore the behavior of self-training over multiple refinement stages. We show that iterative self-training leads to performance degradation if done na\"ively with a fixed ratio of human-labeled to pseudo-labeled training examples. We propose Greedy Iterative Self-Training (GIST) and Random Iterative Self-Training (RIST) strategies that
alternate between training on either human-labeled data or pseudo-labeled data at each refinement stage, resulting in a performance boost rather than degradation.
We further show that GIST and RIST can be combined with existing semi-supervised learning methods to boost performance. 

\end{abstract}


%
\IEEEpeerreviewmaketitle

\section{Introduction}

Semantic segmentation is the task of producing pixel-wise semantic labels over a given image.
This is an important problem that has many useful applications such as medical imaging, robotics, scene-understanding, and autonomous driving. Supervised semantic segmentation models are effective, but they require tremendous amounts of pixel-wise labels, typically provided by a time consuming human annotation process. To overcome the need of collecting more pixel-wise labeled data, there has been
an increase in interest in
semi-supervised semantic segmentation in recent years~\cite{souly2017semi,hung2019adversarial, mittal2019semi,mendelsemi,french2019semi,olsson2020classmix,Ouali_2020_CVPR,alonso2021semi}.

Self-training is a classic semi-supervised learning method that uses pseudo-labels to guide its learning process. We define pseudo-labels as predictions generated by a given model in contrast to human-provided annotations. Self-training means using a model's own predictions as pseudo-labels in its loss during training.
Recently, there has been a resurgence of self-training methods in semi-supervised learning~\cite{lee2013pseudo,radosavovic2018data,zhai2019s4l,zoph2020rethinking,xie2020self}.
Despite the recent comeback of self-training methods, most recent self-training works are confined to only one refinement stage.

Iterative self-training consists of multiple refinement stages, each consisting of K training iterations. At the beginning of each stage the model is initialized with weights from the previous stage, and pseudo-labels are regenerated (see Sec.~\ref{sec2}). We aim to investigate the behaviour of self-training after many (i.e.~$>3$) refinement stages for semi-supervised semantic segmentation.


\begin{figure}[htb]
  \centering
  \includegraphics[width=3.30in, ]{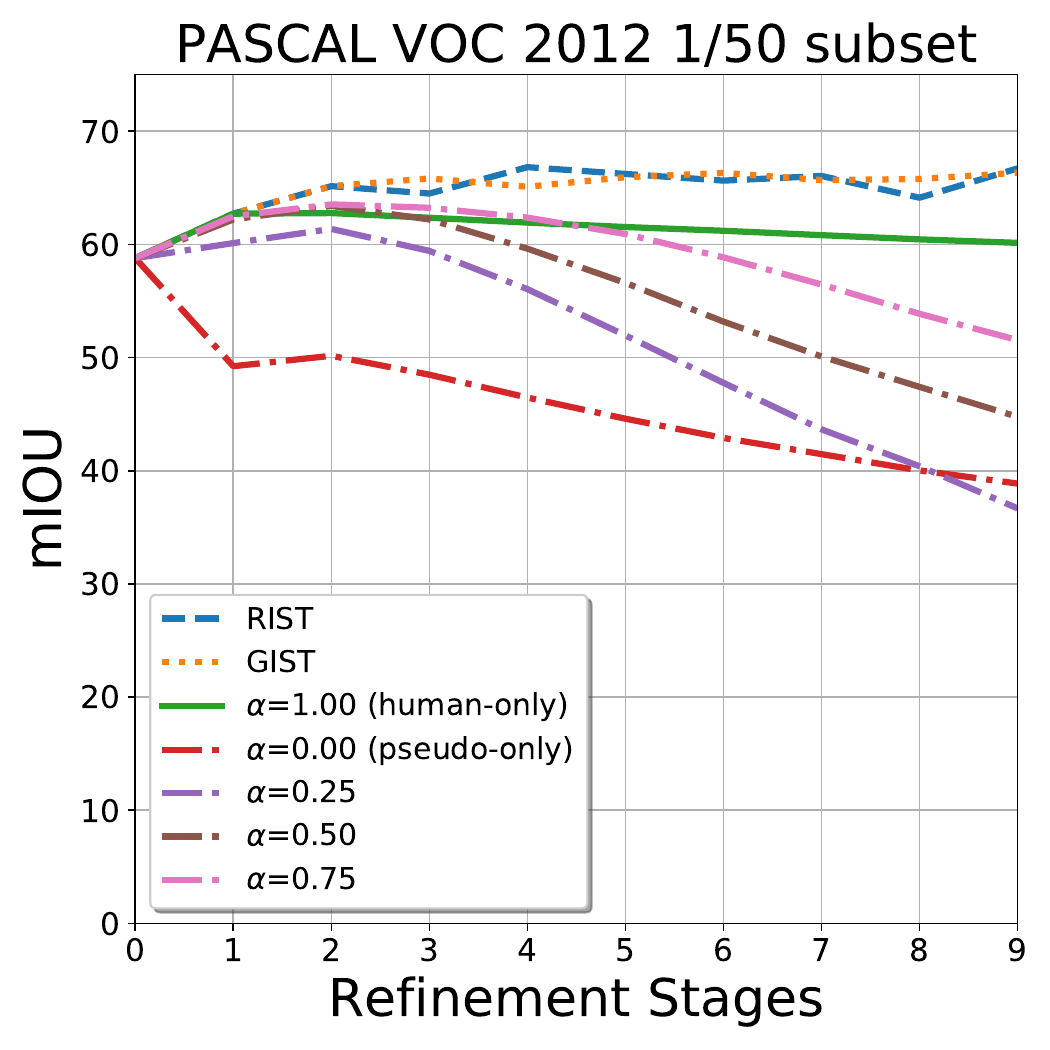}
  \caption{
  The performance of iterative self-training with various ratios of human-labels to pseudo-labels $\alpha$ on the PASCAL VOC 2012 validation datasets. Models are refined iteratively by bootstrapping on weights trained on a previous refinement stage, with only $2\%$ of human-labels. A development set is used to select the best refinement stage.\protect\footnote[2]{}
  }
  \label{fig:voc}
\end{figure}

\begin{figure}[htb]
  \centering
  \includegraphics[width=3.30in, ]{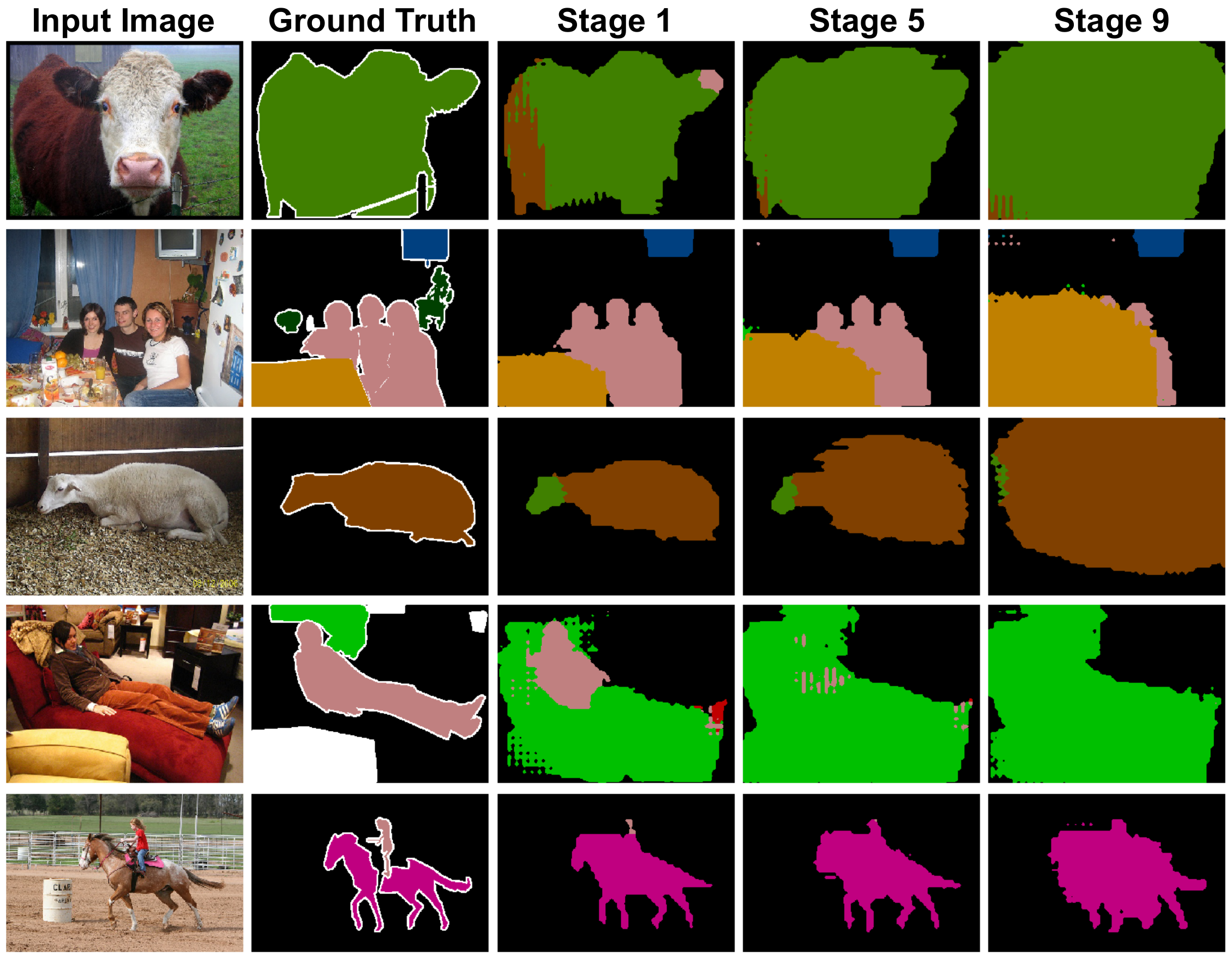}
  \caption{Pseudo-label degradation when a model is trained iteratively with a fixed human-labels to pseudo-labels ratio ($\alpha=0.75$). The first column consists of input images. The second column consists of ground truth labels. The third to fifth columns show pseudo-labels generated after at refinement stage 1, 5 and 9.
  }
  \label{fig:noisy}
\end{figure}

Iterative refinement on a small number of human-labels\footnote{For convenience, we refer to human-labeled training examples as ``human-labels'' and pseudo-labeled training examples as ``pseudo-labels''.} 
may cause over-fitting on the training set, as no new information is introduced. Iterative refinement on pseudo-labels 
does introduce new information which can improve performance. However, it also results in a feedback loop that repeatedly reinforces and compounds incorrect predictions from previous iterations, ultimately resulting in ``pseudo-label bloat'', where a single dominant class prediction spreads to cover an entire image eventually (see Figure~\ref{fig:noisy}).

The na\"ive solution of combining both human-labels and pseudo-labels in each batch slows the rate at which pseudo-label bloat occurs but does not combat it entirely. Instead, we find that alternating training on only human-labels or only pseudo-labels results in a more controlled training dynamic where pseudo-labels help expand predictions to regions that may have been missed, while human-labels prevent pseudo-labels from drifting too far away from the expected annotations. By switching between these two extremes in a greedy fashion (GIST) or random fashion (RIST), our model enjoys the benefits of both label types, ultimately yielding better performance (see Figure~\ref{fig:voc}). Our contributions are the following:
\begin{itemize}
    \itemsep-0em
    \item We show that na\"ive application of iterative self-training to the problem of semi-supervised segmentation via a fixed human-labels to pseudo-labels ratio results in significant performance degradation when $\alpha < 1$.
    \item We introduce Greedy Iterative Self-Training (GIST) and Random Iterative Self-Training (RIST) to overcome performance degradation through iteratively training on either human- or pseudo-labels.
    \item  We demonstrate that both RIST and GIST can improve existing semi-supervised learning methods, yielding performance boost in both the PASCAL VOC 2012 and City- scapes datasets across all eight subsets.
\end{itemize}

\addtocounter{footnote}{1}
\footnotetext{A development set is the set of additional images used for meta-parameter selection (Sec 4 paragraph 1).}\stepcounter{footnote}

\section{Related work}\label{sec:related_work}
 
Semi-supervised semantic segmentation has gained ground in recent years. 
Souly et al.~\cite{souly2017semi} extend a typical Generative Adversarial Network (GAN) network by designing a discriminator that accepts images as input and produces pixel-wise predictions, consisting of confidence maps for each class and a fake label. Hung et al.~\cite{hung2019adversarial} improve GAN-based semantic segmentation by using a segmentation network as a conditional generator.  They also redesign the discriminator to accept the segmentation mask as input and restrict it to produce pixel-wise binary predictions. Mittal et al.~\cite{mittal2019semi} extend Hung et al.'s network by making the discriminator produce an image-level binary prediction. They also add a feature matching loss and self-training loss in their training pipeline. 
They further propose a separate semi-supervised classification network~\cite{tarvainen2017mean} to clean up segmentation masks' prediction.

Recently, a few works propose non-GAN based solutions for semi-supervised semantic segmentation. 
Mendel et al.~\cite{mendelsemi} propose an error-correcting network that aims to fix the predictions of the main segmentation network. This error-correcting network is also applied to unlabeled images to correct the generated mask.
French et al.~\cite{french2019semi} adapt CutMix~\cite{yun2019cutmix} to augment and regularize a segmentation network by creating composite images via mixing two different images and their corresponding segmentation masks.
Similarly, Olsson et al.~\cite{olsson2020classmix} also aim to regularize the segmentation network by mixing segmentation masks from two images by selecting half of the classes from one image and the other half from another image.
Ouali et al.~\cite{Ouali_2020_CVPR} minimize consistency loss between features of multiple auxiliary decoders, in which the perturbed encoder outputs are fed.
Alonso et al.~\cite{alonso2021semi} improve the supervision signals from their teacher model by storing more samples in a separate memory banks.

We explore the iterative self-training method to tackle semi-supervised semantic segmentation. 
Self-training or pseudo-labeling is a classic semi-supervised learning recipe that can be traced back to 1996, where it was used in an NLP application~\cite{yarowsky1995unsupervised}. A major benefit of self-training is that it allows easy extension from an existing supervised model without discarding any information.
In the deep learning literature, Lee et al.~\cite{lee2013pseudo} popularized self-training in semi-supervised classification. After its reappearance, it has gained traction in recent years. Zhai et al.~\cite{zhai2019s4l} show the effectiveness of mixing self-supervised and semi-supervised learning along with pseudo-labeling in their training regime. Zoph et al.~\cite{zoph2020rethinking} show that a randomly initialized model with self-training via a joint-loss can yield better performance than a model initialized with a pre-trained model without self-training. Xie et al.~\cite{xie2020self} show that iterative self-training with noisy labels improves a classification model's accuracy and robustness.
Radosavovic et al.~\cite{radosavovic2018data} use self-training in Omni-supervised learning, where they generate pseudo-labels by taking the average prediction of multiple perturbations of a single unlabeled image.
Most of the recent self-training works~\cite{lee2013pseudo,radosavovic2018data,zhai2019s4l,zoph2020rethinking} are confined to only one refinement stage, 
with the exception of
Xie et al.'s work~\cite{xie2020self}, where they benefit from iterative self-training by repeating self-training for three stages of refinement.
In this work, we aim to extend self-training for semi-supervised semantic segmentation and investigate self-training behavior under many (i.e.~$>3$) refinement stages.\looseness=-1

\section{Methods}\label{sec:methods}

Self-training is a semi-supervised learning method that uses pseudo-labels to guide its learning process. As we improve the model, we also improve the quality of pseudo-labels. Self-training typically consists of the following steps: (1)~training a model using human-labeled data; (2)~generating pseudo-labels using the trained model; and (3)~finetuning the trained model with the combination of human-labeled data and pseudo-labels.

\subsection{Combining both human-labels and pseudo-labels in training}\label{sec1}

Given a set of images $\{(x_i, y_i)\}$, where $x_i$ represents the image and $y_i$ represents the corresponding human-label, we begin training a segmentation model, $\text{SEG}$ for $K$ iterations using a 2D Cross-Entropy Loss, \text{ENT} on available human-labeled data. The loss for iteration $j$ is computed as:
\begin{align}
 L_j & = \frac{1}{B}\sum_{i}^{B}\text{ENT}(o_i, y_i), \label{eq:1} \\
 o_i & = \text{SEG}(x_i).\ \nonumber
\end{align}
\noindent where $B$ represents the batch size, $i$ indexes examples in a batch, and $o_i$ represents a model's output.

After a model is trained on available human-labeled data, we can now use this pre-trained model to generate pseudo-labels on unlabeled data.
Given another set of images $\{(x_i^p, y_i^p)\}$, where $x_i^p$ represents the unlabeled image and $y_i^p$ represents the corresponding pseudo-label, we can now combine human-labels and pseudo-labels by a linear combination of respective losses computed by Eq.~\ref{eq:1}.
\begin{equation}\label{eq:3}
 L_j^{\alpha} = \frac{1}{B}\sum_{i}^{B}(\text{ENT}(o_i, y_i) * \alpha + \text{ENT}(o_i^p, y^p_i) * (1 - \alpha)) .\
\end{equation}

We define $\alpha$ as the ratio of human-labels to pseudo-labels.
Eq.~\ref{eq:1} is a small modification to the classic self-training loss by Lee et al.~\cite{lee2013pseudo},
who apply a coefficient only to the unlabeled loss term, where that coefficient is iteration-dependent.

\subsection{Iterative Self-training}\label{sec2}

Self-training can be repeated through multiple refinement stages, where each refinement stage consists of a pseudo-label generation step and a finetuning step. A na\"ive solution is to fix $\alpha$ throughout all refinement stages,
which we call
Fixed Iterative Self-Training (FIST).
Given that $\alpha$ can take a floating-point number between zero and one, there are infinitely many possible $\alpha$ values at each stage of refinement. By making $\alpha$ binary, we turn the problem into discrete path selection. In this setting, our goal is to find the sequence of stages which yields the best solution. We explore two different selection strategies:  Greedy Iterative Self-Training (GIST) and Random Iterative Self-Training (RIST).


We define $S$ as the maximum number of refinement stages and $K$ as the maximum number of training iterations at a given stage. The number of possible paths in the search space ($2^S$) is exponential in the number of refinement stages.

 \begin{algorithm}[h]
   \caption{GIST}
         \label{alg:GIST}
    \begin{algorithmic}[1]
    \scriptsize
    \STATE $\theta_{list} \xleftarrow{} [\theta_{0}]; \hspace{5pt} \alpha_{list} \xleftarrow{} [0,1]$
    \FOR{$ s \xleftarrow{} 1 \ldots S$ }
        \STATE $\theta_{list}^{*} \xleftarrow{} [\ ]; \hspace{5pt} \text{R}_{list}^{} \xleftarrow{} [\ ]$
        \FOR{$\theta_c \ \ \text{in} \ \ \theta_{list}$}
            \FOR{$\alpha \ \ \text{in} \ \ \alpha_{list}$}
                \FOR{$m \xleftarrow{} 1 \ldots $ M}
                    \STATE $y^p \xleftarrow{} \argmax(SEG(x_m^p))$
                \ENDFOR

                \FOR{$j \xleftarrow{} 1 \ldots $ K}
                    \STATE $ L_j^{\alpha} \xleftarrow{} \frac{1}{B}\sum_i^B(\text{ENT}\left(o_i, y_i\right) * \alpha  +
                     \text{ENT}\left( o_i^p, y^p_i\right) * (1 - \alpha))$
                    \STATE $\theta_{c} \xleftarrow{} \theta_{c} - \lambda \frac{\partial L_j^{\alpha}}{\partial \theta_{c}}$
                \ENDFOR
                \STATE append $\theta_c$ to $\theta_{list}^{*}$
                \STATE append $\text{EVAL}(\theta_c)$ to $R_{list}$
            \ENDFOR
        \ENDFOR
        \STATE sort $\theta_{list}^{*}$ based on $R_{list}$ in descending order
        \STATE keep top $G$ items in $\theta_{list}^{*}$ and assign it to $\theta_{list}$
    \ENDFOR
\end{algorithmic}
\end{algorithm}

\begin{algorithm}[h]
   \caption{RIST}
         \label{alg:RIST}
    \begin{algorithmic}[1]
    \scriptsize
    \FOR{$ s \xleftarrow{} 1 \ldots S$ }
        \STATE $\alpha \xleftarrow{} \left(\text{Uniform}(0,1) > 0.5\right)$
        \STATE $\theta_{s} \xleftarrow{} \theta_{s-1}$
        \FOR{$m \xleftarrow{} 1 \ldots $ M}
            \STATE $y^p \xleftarrow{} \argmax(SEG(x_m^p))$
        \ENDFOR

        \FOR{$j \xleftarrow{} 1 \ldots $ K}
            \STATE $ L_j^{\alpha} \xleftarrow{} \frac{1}{B}\sum_i^B(\text{ENT}\left(o_i, y_i\right) * \alpha  +   \text{ENT}\left( o_i^p, y^p_i\right) * (1 - \alpha))$
            \STATE $\theta_{s} \xleftarrow{} \theta_{s} - \lambda \frac{\partial L_j^{\alpha}}{\partial \theta_{s}}$
        \ENDFOR
    \ENDFOR
\end{algorithmic}
\end{algorithm}

\begin{figure}[htb]
  \centering
  \includegraphics[width=3.30in, ]{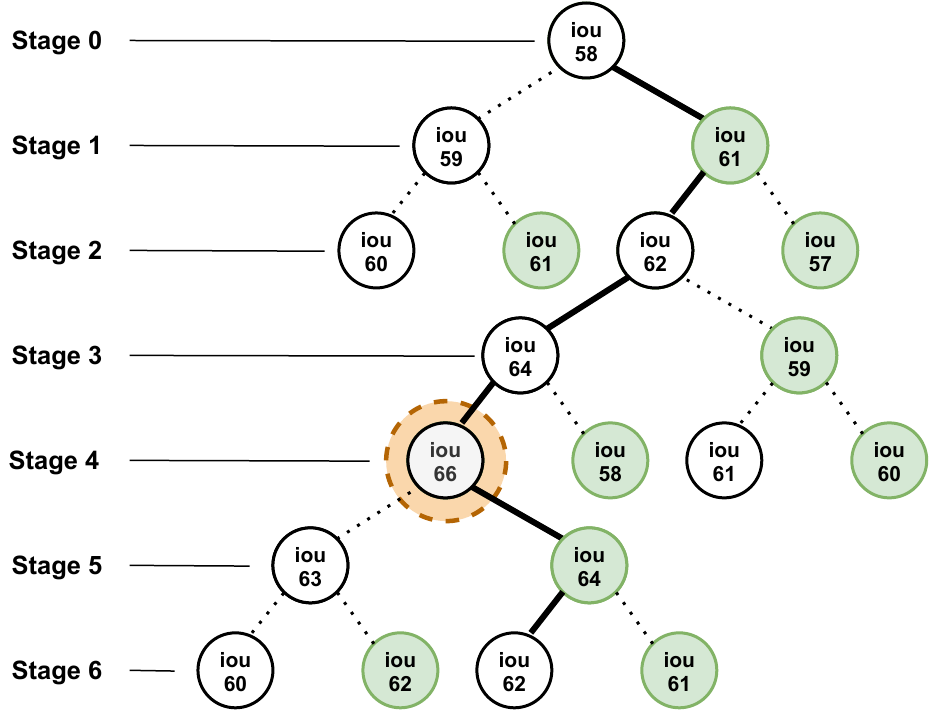}
  \caption{A hypothetical $\alpha$ path selection scenario with six refinement stages in self-training using the greedy approach (GIST). $\alpha$ indicates the ratio of human-labels to pseudo-labels. The open nodes indicate that only human-labels ($\alpha = 1$) are being used for training and the shaded nodes indicate only pseudo-labels ($\alpha = 0$) are being used. The number of possible paths is exponential in the number of refinement stages ($2^6$).
  At each stage of refinement, we evaluate the mean intersection over union (mIOU) of a model using a development set. Here, the optimum value is found at stage four of the refinement process.
  }
  \label{fig:sel}
\end{figure}

GIST works by finding the best stage of refinement by relying on the development set.
Algorithm~\ref{alg:GIST} describes the GIST algorithm, which is a beam search strategy. In line 3, we keep a list of candidates $\theta$, and the corresponding evaluation results, $R_{list}$, using the $\text{EVAL}$ function on the development set. These lists are updated in lines 13 and 14. In line 17, we sort $\theta^*_{list}$ in descending order based on $R_{list}$. In line 18, we keep the top G $\theta$ in the $\theta^*_{list}$. and assign it to $\theta_{list}$, where it will be used as the initial model for the next stage of refinement.
In Line 6 to 8, we generate our pseudo-labels for $M$ unlabeled images. In Line 9 to 12, we finetune a segmentation model for $K$ iterations with learning rate $\lambda$. Figure~\ref{fig:sel} shows a hypothetical $\alpha$ path selection scenario using GIST.

One weakness of GIST is its smaller search space when compared to a random search (RIST). Bergstra et al.~\cite{bergstra2012random} show that in low dimensions, random search is an effective search strategy compared to a grid search. One can see that a greedy search is a subset of a grid search where we only explore the top performing branches. With a beam size of 1, the search space for a greedy search is $\log(S)$, and the search space for a random search is $2^S$. At first glance, RIST may seem counter-intuitive, but RIST's performance spread is relatively small ($\pm$0.93), and this spread can be reduced by eliminating obvious degenerate solutions and increasing search paths (see Sec.~\ref{sec:discussion} for details).
Algorithm~\ref{alg:RIST} describes the RIST algorithm.  In line 2, we randomly set $\alpha$ to either zero or one.  Line 7 to 10 is similar to the GIST algorithm (line 9 to line 12).


In GIST, the time complexity of a beam search is $O(S*G)$. A beam search is only as efficient as a random search if we can parallelize the training for each $\alpha$ at each stage, and this condition requires a beam search to be trained on consecutive numbers of GPU resources. Unlike a beam search, each random search can be trained independently on a single GPU resource. It is easier and cheaper to obtain $N$ independent GPUs rather than $N$ consecutive GPU resources, making RIST computationally cheaper and faster in practical settings.

\subsection{Additional Add-ons}

In addition to cross entropy loss in Eq~\ref{eq:3}, we include three additional add-ons to boost the performance of FIST, GIST and RIST: Consistency Loss (CL), Label Erase (LE), and Temperature Scaling (TS).
Table~\ref{table:addons} shows the ablation study of the add-ons for both RIST and GIST for models trained on the Pascal VOC 1/50 subset. \\

\noindent\textbf{Consistency Loss (CL)}:
Consistency Loss (CL) is a common loss in semi-supervised learning where the goal is to minimize features between two perturbations of the same input. 
CL is a common loss in semi-supervised learning~\cite{mittal2019semi,olsson2020classmix}. Mittal et al.~\cite{mittal2019semi} use CL via a feature matching loss to minimize the discrepancy between predicted features and ground truth features. Olsson et al.~\cite{mittal2019semi} use CL via mean-teacher~\cite{tarvainen2017mean} method. We also use the mean-teacher method as our CL loss.
We use the following equations (Equations~\ref{eq:mt} to~\ref{eq:f2}) to calculate CL for both $x_i$ and $x_i^p$. 

\setcounter{equation}{2}
\begin{align}
 & \theta^* = \theta^ * \beta + \theta * (1 - \beta) )\label{eq:mt}\\
 & o_i,f_i = \text{SEG}(x_i)\label{eq:x1}\\
 & o_i^*,f_i^* = \text{SEG}^*(x_i)\label{eq:x2}\\
 & f_i = \text{drop}(\text{pool}(f_i))\label{eq:f1}\\
 & f_i^* = \text{drop}(\text{pool}(f_i^*))\label{eq:f2}\\
 &  L_{\text{feature}} = | f_i - f_i^*|
\end{align}

\begin{figure}[htb]
  \centering
  \includegraphics[width=3.00in, ]{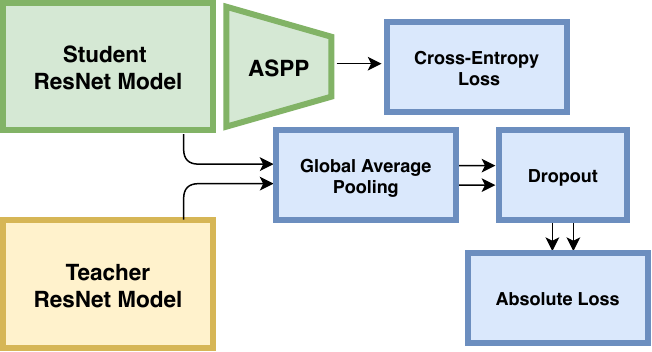}
  \caption{Depiction of the Consistency Loss (CL). At each batch of training, there are two copies of Resnet Models. The student model is the main model that is used to predict the segmentation mask, and the teacher model is a copy of the student model with an exponential moving weight update.  The goal of CL is to minimize the differences between the features of the student and teacher models.}
  \label{fig:cl}
\end{figure}

Following ~\cite{mittal2019semi,hung2019adversarial}, we use DeepLabV2~\cite{chen2017deeplab} as our segmentation model; therefore, $f_i$ represents features before an Atrous Spatial Pyramid Pooling (ASPP) layer. Equation~\ref{eq:mt} describes the weight update rule for the teacher model, $\text{SEG}^*$, where an exponential moving average rule is applied to its weights, $\theta^*$, which is controlled by $\beta$, $(0 \leq \beta \leq 1)$.
Equations~\ref{eq:x1} and \ref{eq:x2} illustrate the extraction of features before the ASPP layer in both the student model, $\text{SEG}$, and the teacher model, $\text{SEG*}$. In Equations~\ref{eq:f1} and \ref{eq:f2}, we apply global average pooling followed by a dropout perturbation to these features. After that, we compute the absolute differences between these two features.
Figure~\ref{fig:cl} summarizes the feature consistent loss in our model. \\

\noindent\textbf{Label Erase (LE)}:
Pseudo-labels are often very noisy, especially when we have a small amount of human-labeled data.
To remove noise from pseudo-labels, we only keep predictions that pass a certain confidence threshold, $\phi$. 
Label Erase (LE) is also used in \cite{mittal2019semi,olsson2020classmix,radosavovic2018data,zoph2020rethinking}. 


Equation~\ref{eq:conf} illustrates the process of flagging low confidence prediction regions so that they are ignored by the loss function in Equation 1. We define pixel-wise confidence, $c_{i,j}^p$ as the pixel softmax output of $o_{i,j}^p$, where $i$ represents the image index and $j$ represents the pixel index.


\begin{align}\label{eq:conf}
        y_{i,j}^p = \begin{cases}
            \argmax(c_{i,j}^p), & \text{if}\; \max(c_{i,j}^p) >= \phi,\\
            \text{ignore label}, & \text{if}\; \max(c_{i,j}^p) < \phi,
        \end{cases}
\end{align} \\

\begin{figure}[htb]
  \centering
  \includegraphics[width=3.28in, ]{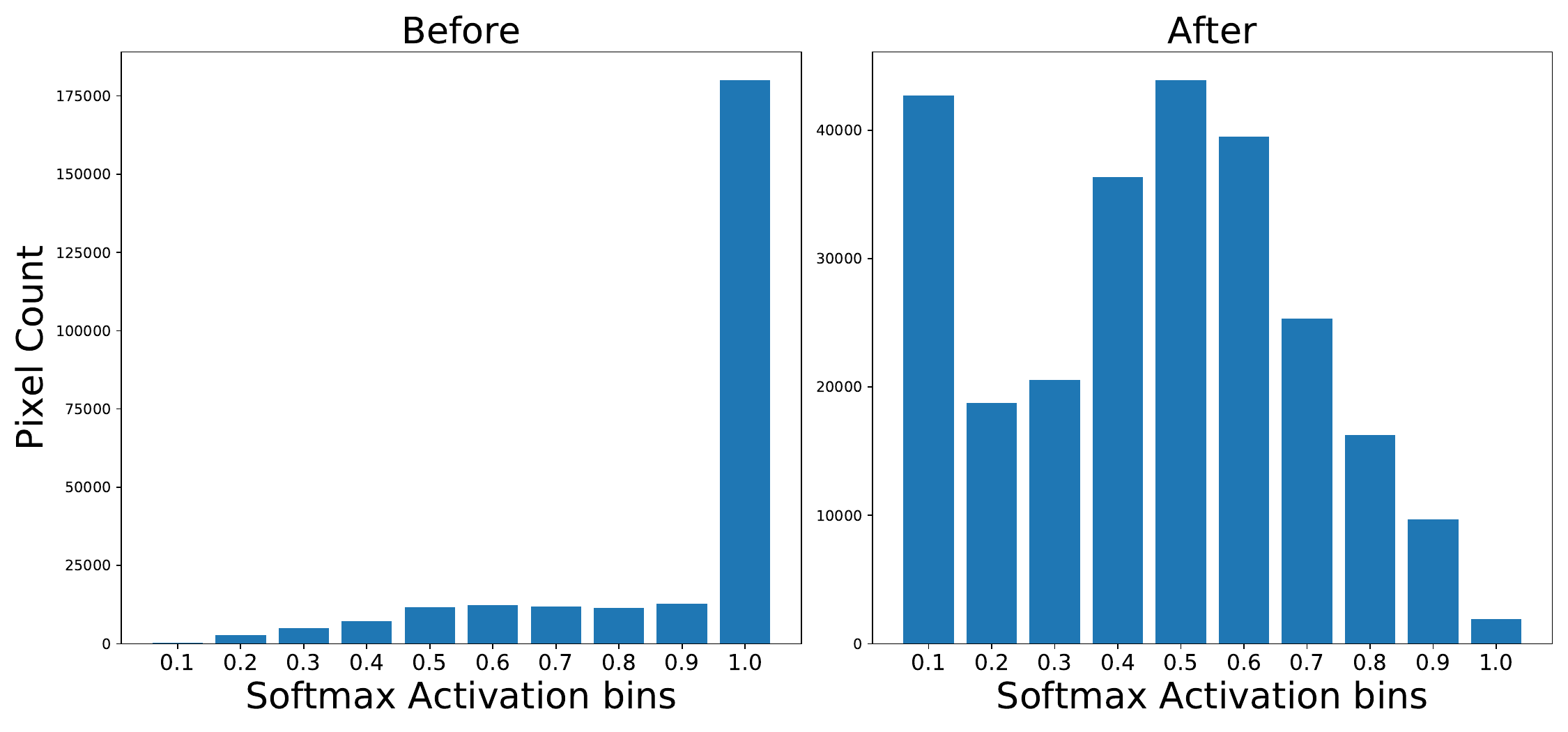}
  \caption{The before and after effects of applying temperature scaling to the output activation of a single image. A temperature scale of 0.2 is applied to the output activation.
  }
  \label{fig:ts}
\end{figure}

\noindent\textbf{Temperature Scaling (TS)}:
Temperature scaling (TS) is introduced by Hinton et al.~\cite{hinton2015distilling}, where it is used to create a softer probability distribution for knowledge distillation. The formula for TS is defined as 
$q_{i} = \frac{\exp(y_i*T)}{\sum_j \exp(y_j*T)}$. As $T$ becomes smaller, the output of the softmax function will tend towards uniform distribution. 
We employ TS~\cite{hinton2015distilling} to overcome over-confident prediction in our model, where the activation values after softmax are highly skewed towards $100\%$ (see Figure~\ref{fig:ts}).

\section{Experiments}\label{sec:experiment}

\begin{table*}[!htb]
\centering
\scalebox{1.0}{
\setlength{\tabcolsep}{15pt}
\begin{tabular}{|c||c|c|c||c|c|c|c|c|}
\hline
&\multicolumn{3}{c||}{VOC 2012 ($\approx$10k images)}&\multicolumn{5}{c|}{Cityscapes ($\approx$3k images)} \\ \hline \hline
$\#$ of labeled images & 211 & 529 & 1,322 & 59 & 100 & 148 & 371 & 743  \\ \hline
Subset & 1/50 & 1/20 & 1/8 & 1/50 & 1/30 & 1/20 & 1/8 & 1/4   \\ \hline
Supervised & 54.15 & 62.94 & 67.44 & 49.68 & 53.96 & 54.71 & 59.90 & 62.21 \\
Supervised + GIST & 66.33 & 66.95 & 70.27 & \textbf{53.51} & 56.38 & 58.11 & 60.94 & 63.04 \\
Supervised + RIST & 66.71 & 68.28 & 69.90 &  53.33 & 56.28 & 57.81 & 61.38 & 63.92 \\

\hline
S4GAN~\cite{mittal2019semi} & 62.87 & 62.35 & 68.56 & 50.48 & 54.58 & 55.61 & 60.95 & 61.30 \\
S4GAN~\cite{mittal2019semi} + GIST & \textbf{67.21} & 68.50 & 70.61 & 52.36 & 57.18 & 57.40 & 61.27 & 64.24 \\
S4GAN~\cite{mittal2019semi} + RIST & 66.51 & 68.50 & 70.31 & 53.47 & 57.12 & 57.48 & 62.50 & 64.64 \\
\hline
ClassMix~\cite{olsson2020classmix} & 63.63 & 66.74 & 66.14 & 52.14 & 57.02 & 58.77 & 61.56 & 63.90 \\
ClassMix~\cite{olsson2020classmix} + GIST & 65.60 & 69.05 & 70.65 & 52.43 & \textbf{58.70} &  \textbf{59.98} & 62.44 & 64.53 \\
ClassMix~\cite{olsson2020classmix} + RIST & 66.30 & \textbf{69.40} & \textbf{70.76} & 53.05 & 58.55 & 59.54 & \textbf{62.57} & \textbf{65.14} \\
\hline 
\end{tabular}
}
\bigskip
\caption{Semantic segmentation results (mIoU) on the PASCAL VOC 2012 and Cityscapes validation datasets.}
\label{table:det}
\end{table*}

We perform experiments on two semantic segmentation datasets: PASCAL VOC 2012~\cite{pascal-voc-2012} and Cityscapes~\cite{Cordts2016Cityscapes}. In each dataset, there are three subsets, where pre-defined ratios ($1/50$, $1/20$, and $1/8$) of training images are selected as images with human-labels. We also experiment on two additional subsets ($1/30$ and $1/4$) for the Cityscapes dataset. These subsets of labeled images are selected using the same split as ~\cite{hung2019adversarial,mittal2019semi,olsson2020classmix}. We treat the remaining images as unlabeled examples.  We use the mean intersection-over-union (mIoU) as a performance metric. The validation images for both datasets are set aside and used for evaluation, which is consistent with~\cite{hung2019adversarial, mittal2019semi,olsson2020classmix}. We select 50 additional images from the training set as a modest ``development set'' for meta-parameter tuning.

For all of our experiments, we follow the same experimental setup as \cite{hung2019adversarial,mittal2019semi,olsson2020classmix}. We use a DeepLabV2~\cite{chen2017deeplab} segmentation model that is initialized with MS-COCO pre-trained weights~\cite{lin2014microsoft}~\footnote[5]{DeepLabV2 backbone is used in our experiments so that our method is comparable to S4GAN and ClassMix. }. We also freeze all the BatchNorm Layers in DeepLabV2. We optimize our model using the Stochastic Gradient Descent (SGD) optimizer with a base learning rate of 2.5e-4, a momentum of 0.9, and a weight decay of 5e-4. We use a polynomial learning rate decay policy, where we adjust the learning rate with the following equation:
$\lambda_{\text{iter}} = \lambda_0 (1- \frac{\text{iter}}{\text{max\_iter}})^{0.9}$ where $\lambda_0$ is a base learning rate. To augment the dataset, we use random-cropping ($321\times321$ for PASCAL VOC 2012 and $256\times512$ for Cityscapes), horizontal-flipping (with a probability of 0.5), and random-resizing (with a range of 0.5 to 1.5) in all of our experiments.

For all subsets, we train our supervised models and stage-0 models for 25,000 iterations. 
Additionally, for the Pascal VOC 2012 dataset, we use a batch-size of 8, and we refine our model for 3,000 iterations at each refinement stage (Stage 1 to 9). For the Cityscapes dataset, we use a batch-size of 6, and we refine our models for 4,000 iterations at each refinement stage (Stage 1 to 9). We use a search cost of two for both GIST and RIST and use the development set to select the best path for all experiments.

Table~\ref{table:det} shows the results of our experiments as well as relevant
results that others have reported~\cite{hung2019adversarial,mittal2019semi,french2019semi,olsson2020classmix}. We use the code provided by the respective authors for our experiments with S4GAN~\cite{mittal2019semi} and ClassMix \cite{olsson2020classmix}. For a fair comparison, we set the batch size for S4GAN and ClassMix to match with our experiments, and we also select the best iterations
using our development set.
We notice performance differences in S4GAN and ClassMix when compared to performances reported in the original papers. We speculate that the differences are caused by best iteration selection and batch sizes.
For experiment with S4GAN+GIST/RIST and ClassMix+GIST/RIST, we 
first train the segmentation model with S4GAN and ClassMix algorithm (stage-0 models). After that, we further refine the segmentation model using the GIST/RIST algorithm by bootstrapping on the DeepLabV2 model trained with S4GAN/ClassMix.


\subsection{Discussion}\label{sec:discussion}

Figure~\ref{fig:voc} shows that a na\"ive application of iterative self-training leads to significant performance degradation in both datasets. Figure~\ref{fig:noisy} shows the qualitative evidence of performance degradation in FIST at $\alpha=0.75$. FIST suffers from over-confident pseudo-label predictions which spread to the surrounding pixel. Over multiple stages of refinement, the pseudo-labels expand and eventually engulf most of the image. We speculate that this may be why most of the recent self-training works are confined to one refinement stage.

\begin{figure*}[!htb]
  \centering
  \includegraphics[width=6.50in, ]{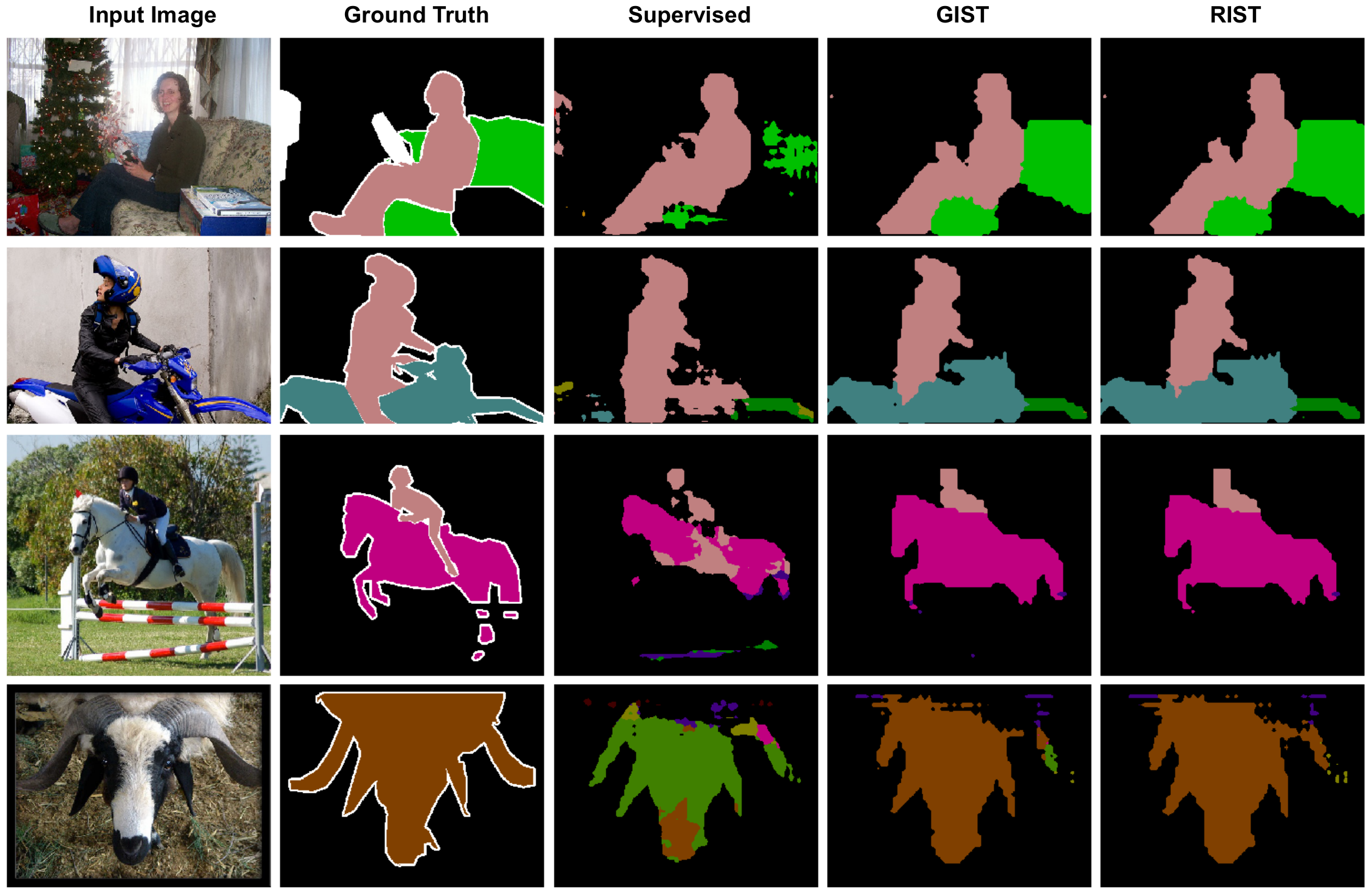}
  \caption{The qualitative results of our model train on $2\%$ human-labels from the PASCAL VOC 2012 dataset.}
  \label{fig:voc_vis}
\end{figure*}

\begin{figure*}[!htb]
  \centering
  \includegraphics[width=6.50in, height=3.50in,]{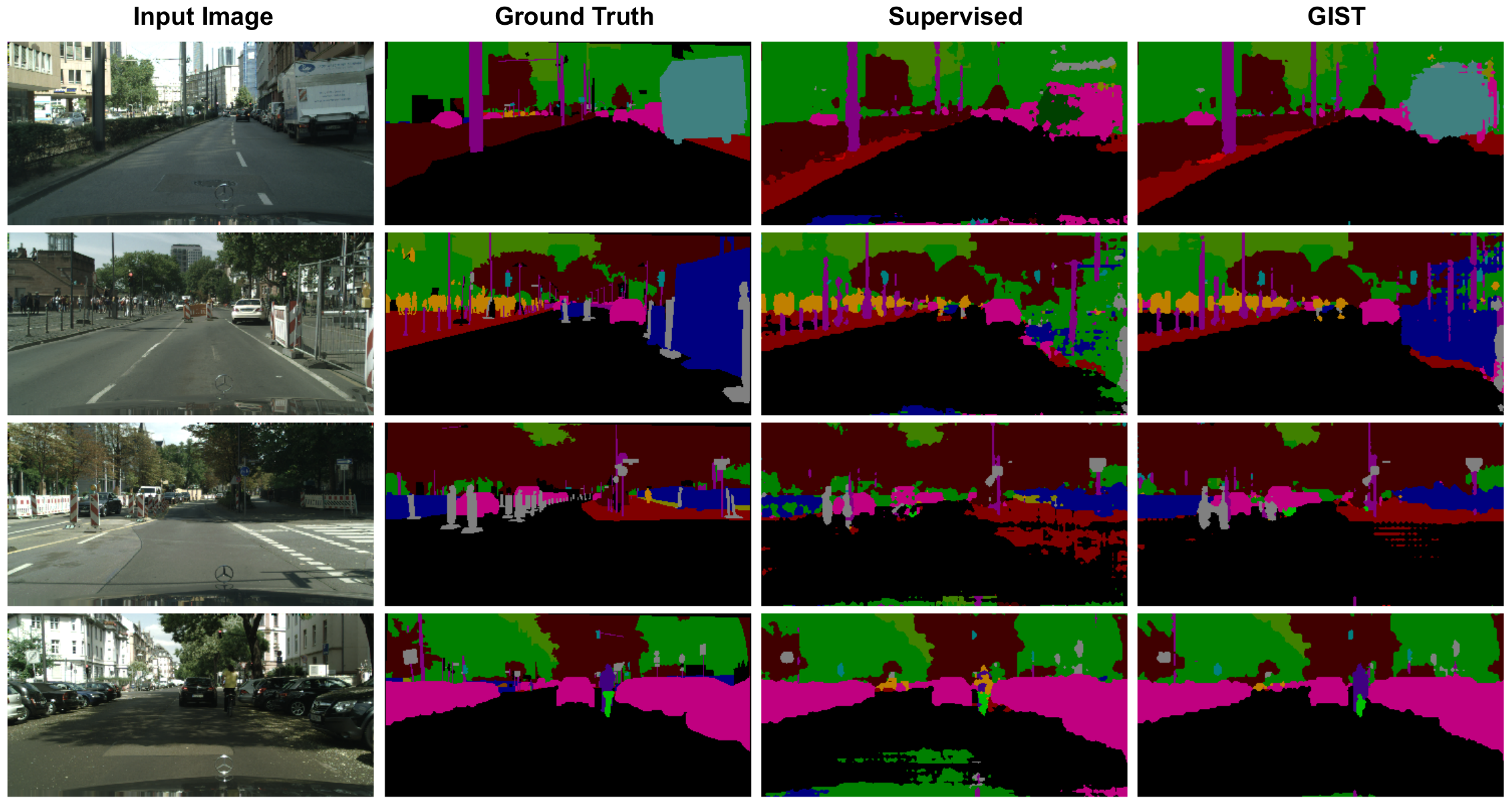}
  \caption{The qualitative results of our model trained on a $2\%$ human-labels from the Cityscapes dataset.}
  \label{fig:city_vis}
\end{figure*}

Figure~\ref{fig:voc} also shows that both RIST and GIST overcome performance degradation. Additionally, both RIST and GIST generalize better than FIST in each successive refinement.
Table~\ref{table:det} shows that both RIST and GIST can improve other semi-supervised segmentation techniques such as S4GAN and ClassMix. 
We show that both RIST and GIST can further refine S4GAN and ClassMix yielding performance boosts across all subsets in the Pascal VOC 2012 and Cityscapes datasets.
Figures~\ref{fig:voc_vis} and~\ref{fig:city_vis} show RIST and GIST's qualitative results that are trained with $2\%$ of human-labels in both datasets.

\begin{figure}[!htb]
  \centering
  \includegraphics[width=3.30in, ]{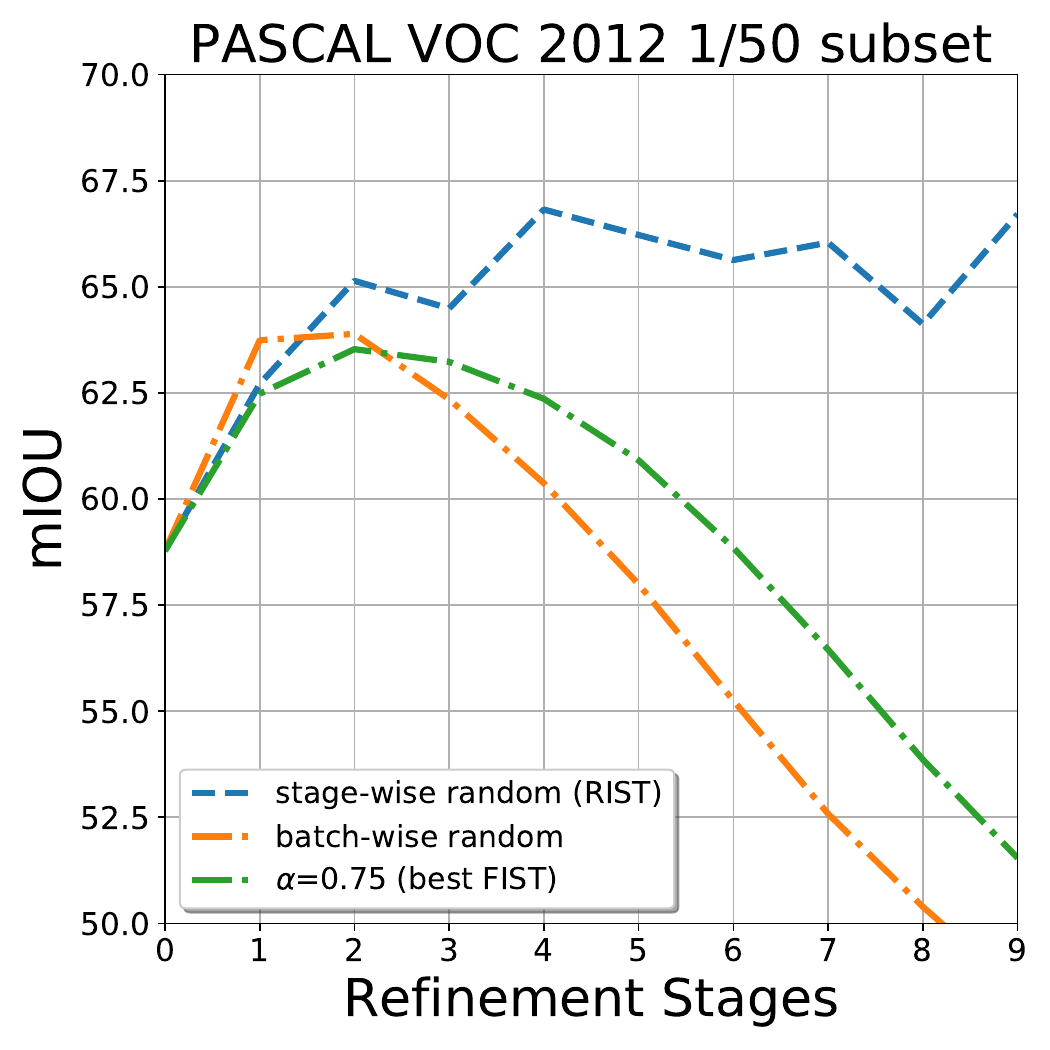}
  \caption{Self-training performance at various stages between batch-wise and stage-wise random selection strategies on the PASCAL VOC 2012 validation set.}
  \label{fig:batch}
\end{figure}

Figure~\ref{fig:batch} demonstrates the importance of training on a single label type (i.e., human-label or pseudo-label) for an extended number of iterations. We find that randomly selecting the label type for each batch (batch-wise random) performs just as poorly as training with both label types in each batch (FIST). We speculate that the clean human-labels and the noisy pseudo-labels represent competing objectives, which are difficult for the model to satisfy simultaneously, resulting in it getting stuck at poor solutions. By applying stage-wise training, we allow the model to focus on a single objective at a time, potentially escaping sub-optimal solutions from a previous stage.

\begin{table}[!htb]
\centering
\scalebox{1.0}{
\setlength{\tabcolsep}{2pt}
\begin{tabular}{|l|c|c|c|}
\hline
Selection & mIoU (devel) & mIoU (val) & Group mIoU (val) \\ \hline
PPLLPLPLL & 54.35 & 66.14 & 67.03 \\
LPPLLLPPL & 55.05 & 66.71 &   \\
PPLLPPPLP & 54.21 & 66.05 &   \\
LPLPPLLPL & 55.12 & 67.03 &   \\
LLLLLLPPP & 54.43 & 63.75 &  \\
\hline
PPPPPPLLP & 50.43 & 64.37 & 66.63  \\
PPLLLPLPL & 54.00 & 66.02 &   \\
LLPLPLLPL & 55.19 & 66.24 &   \\
LPLPLLPLP & 55.29 & 66.63 &   \\
LLLPLPPLP & 54.88 & 65.17 &   \\
\hline
PLPLPPPLP & 54.64 & 66.40 & 66.62 \\
LPPLPPLPP & 54.46 & 66.83 &   \\
LPPLLLLPP & 54.67 & 65.61 &   \\
LPPPLLLLP & 54.82 & 66.62 &  \\
LPPPLLPPL & 54.64 & 66.61 &   \\
\hline
Mean$\pm$1 std.~dev.&  & 66.01$\pm$0.93 & 66.76$\pm$0.23 \\
\hline
\end{tabular}
}
\bigskip
\caption{RIST semantic segmentation results on the PASCAL VOC 2012 validation set. Group experiment results are selected based on the best development mIoU. Random selection choices are P (pseudo-label only) or L (human-label only). There are a total of nine refinement stages ordered sequentially from left to right. }
\label{table:stable}
\end{table}

\begin{figure}[!htb]
  \centering
  \includegraphics[width=3.30in, ]{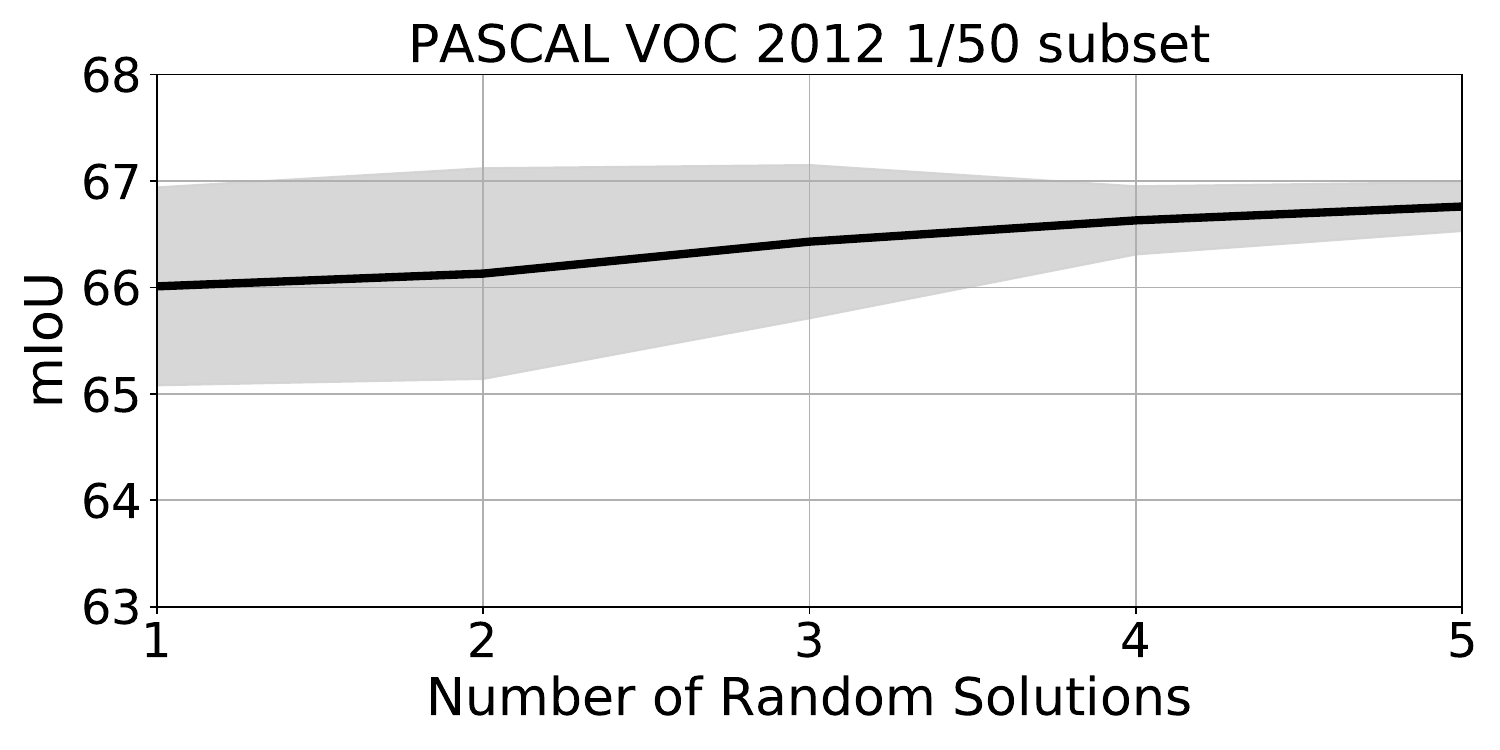}
  \caption{The effect of increasing stability of RIST when we increase the number of random solution. The best results of the random solutions are selected based on best development mIoU. The shaded area represent one standard deviation of uncertainty.
  }
  \label{fig:stable}
\end{figure}

Table~\ref{table:stable} examines the stability of performance for RIST on the PASCAL VOC 2012 1/50 subset. We use the same subset for training and generate fifteen different permutations of binary $\alpha$ values uniformly at random.
The degenerate solutions occur in row 5 and 6, where we have more than four consecutive numbers of the same $\alpha$ choice during training. If we were to perform a random search once, the standard deviation of mIoU is 0.93. If we were to remove the obvious degenerate solutions (row 5 and 6) using some heuristics, the standard deviation of mIoU is reduced to 0.52. Nevertheless, we can reduce this standard deviation further by selecting the best solution out of five different random solutions yielding a standard deviation of 0.23. Figure~\ref{fig:stable} shows that as we increase the number of random solutions, the spread decreases.

\begin{table}[!htb]
\centering
\setlength{\tabcolsep}{13pt}
\scalebox{1.0}{
\begin{tabular}{|l|c|c|}
\hline
Sample Size & mIoU (RIST) & mIoU (GIST) \\ \hline
10 & 66.04 & 66.23\\
25 & 65.63 & 62.76\\
50 & 66.71 & 66.33\\
100 & 66.71 & 66.67 \\
200 & 66.71 & 66.69\\
500 & 66.71 & 67.00 \\
\hline
\end{tabular}
}
\bigskip
\caption{Semantic segmentation results on the PASCAL VOC 2012 validation set for RIST and GIST based on best epoch selected while varying the number of examples in the development set.}
\label{table:devel}
\end{table}

\begin{table}[!htb]
\centering
\setlength{\tabcolsep}{5pt}
\scalebox{1.0}{
\begin{tabular}{|p{1cm}|p{1cm}|p{1cm}|p{1cm}|c|}
\hline
Beam Size & Search Cost & mIoU (devel) & mIoU (val) & Solution \\ \hline
1 & 2 & 55.17 & 66.33 & LPLLPLLPL\\
2 & 4 & 55.46 & 64.95 & LLPLLPLLP\\
3 & 6 & 55.46 & 64.95 & LLPLLPLLP\\
\hline
\end{tabular}
}
\bigskip
\caption{Semantic segmentation results on the PASCAL VOC 2012 validation set for GIST on different beam size. selection choices at each stage are P (pseudo-label only) or L (human-label only).}
\label{table:gist_beam}
\end{table}

Table~\ref{table:devel} explores the sensitivity of RIST and GIST's performance on the PASCAL VOC 2012 $1/50$ subset with respect to the size of the development set. In general, RIST and GIST are relatively stable. RIST performance remains the same for 50 to 500 sample sizes. GIST performance improves slightly as we increase the sample size from 50 to 500.
On the Pascal VOC 2012 dataset, we show that our supervised+GIST and supervised+RIST trained with 211 human-labels ($1/50$ subset) outperform the supervised model that is trained with 529 human-labels ($1/20$ subset). This result shows that GIST and RIST improve the model, not just because they got more supervised signals from the development set (see Table~\ref{table:det}).

Table~\ref{table:gist_beam} explores GIST at various beam sizes. GIST can find a better solution for the development set; however, since there is a mismatch between the distribution of the development set and the original validation set due to the small sample size, the best solution of the development set is not the best solution for the original validation set. This study shows that GIST has a higher chance to overfit the development set when compared to RIST.

\begin{table}[!htb]
\centering
\setlength{\tabcolsep}{15pt}
\begin{tabular}{|l||l|l|}
\hline
Method & RIST & GIST \\
\hline
no add-on & 60.80 & 58.23  \\
+CL  & 61.72 & 62.37 \\
+CL +LE & 64.69 & 64.56\\
+CL +LE +TS & 66.71 & 66.33 \\
\hline
\end{tabular}
\bigskip
\caption{Ablation study of the add-ons for both RIST and GIST for models trained on the Pascal VOC 1/50 subset. Results are reported in mIOU on the validation set.}
\label{table:addons}
\end{table}

\section{Conclusion}

We show that iterative self-training with a fixed human-labels to pseudo-labels ratio (FIST) leads to performance degradation. This degradation can be overcome by alternating training on only human-labels or only pseudo-labels in a greedy (GIST) or random (RIST) fashion. A clear benefit of self-training is that it can easily extend existing architectures. 
We show that both GIST and RIST can further refine models trained with other semi-supervised techniques resulting in a performance boost.

\bibliographystyle{IEEEtran}
\bibliography{crv}

\end{document}